\documentclass[10pt,twocolumn,letterpaper]{article}

\usepackage{iccv}
\usepackage{times}
\usepackage{epsfig}
\usepackage{graphicx}
\usepackage{amsmath}
\usepackage{amssymb}
\usepackage{enumitem}
\usepackage{mathrsfs}
\usepackage{booktabs}
\usepackage[caption=false]{subfig}
\usepackage{multicol}
\usepackage{gensymb}
\usepackage{mathtools, cases}
\usepackage{fancyhdr}

\usepackage[pagebackref=true,breaklinks=true,letterpaper=true,colorlinks,bookmarks=false]{hyperref}

\iccvfinalcopy 

\def\iccvPaperID{****} 

\begin{document}
	
\title{U4D: Unsupervised 4D Dynamic Scene Understanding}

\author{Armin Mustafa  \hspace{.125\linewidth} Chris Russell  \hspace{.125\linewidth} Adrian Hilton\\
	CVSSP, University of Surrey, United Kingdom\\
	{\tt\small \{a.mustafa, c.russell, a.hilton\}@surrey.ac.uk}
}

\maketitle
\thispagestyle{fancy}

\begin{abstract}
	We introduce the first approach to solve the challenging problem of unsupervised 4D visual scene understanding for complex dynamic scenes with multiple interacting people from multi-view video. 
	Our approach simultaneously estimates a detailed model that includes a per-pixel semantically and temporally coherent reconstruction, together with instance-level segmentation exploiting photo-consistency, semantic and motion information. 
	We further leverage recent advances in 3D pose estimation to constrain the joint semantic instance segmentation and 4D temporally coherent reconstruction. This enables per person semantic instance segmentation of multiple interacting people in complex dynamic scenes.
	Extensive evaluation of the joint visual scene understanding framework against state-of-the-art methods on challenging indoor and outdoor sequences demonstrates a significant ($\approx40\%$) improvement in semantic segmentation, reconstruction and scene flow accuracy.
\end{abstract}
\vspace{-0.85cm}
\section{Introduction}
\begin{figure}
	\begin{center}
		\includegraphics[width=0.47\textwidth]{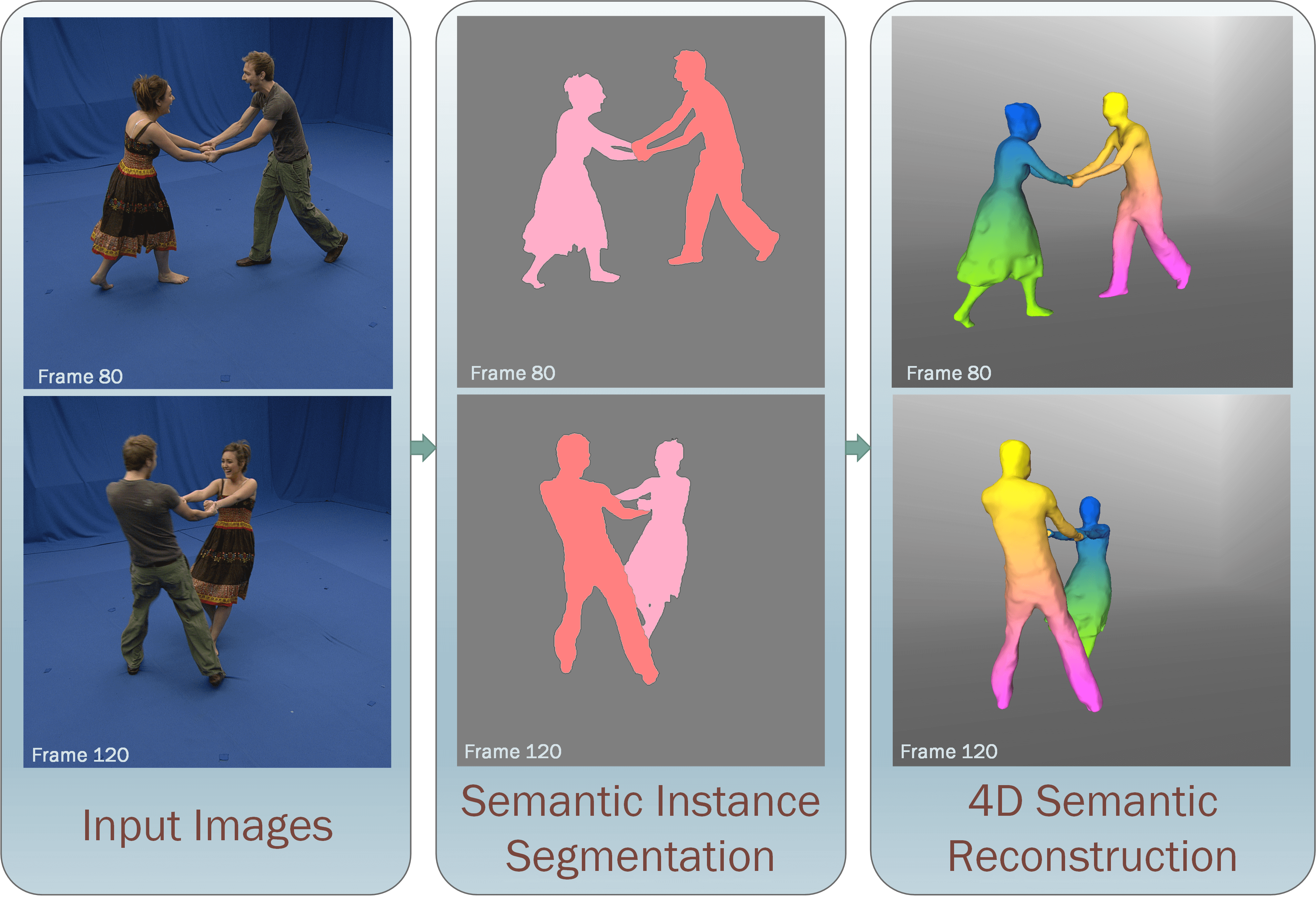}
		\caption{Joint 4D semantic instance segmentation and reconstruction exploiting 3D human-pose of interacting people in dynamic scenes. Shades of pink in  segmentation represents instances of people. Colour assigned to reconstruction of frame 80 is reliably propagated to frame 120 using proposed temporal coherence.}
		\label{fig:motivation}
		\vspace{-0.25cm}
	\end{center}
\end{figure}

With the advent of autonomous vehicles and rising demand for immersive content in augmented and virtual reality, understanding dynamic scenes has become increasingly important.
In this paper we propose an unsupervised framework for 4D dynamic scene understanding to address this demand.
By ``4D Scene understanding'' we refer to a unified framework that describes: 3D modelling; motion/flow estimation; and semantic instance segmentation on a per frame basis for an entire sequence. 
Recent advances in pose estimation \cite{cao2017realtime,Tome_2017_CVPR} and recognition \cite{2017maskrcnn,Xie2016CVPR,ChenPK0Y16} using deep learning have achieved excellent performance for complex images.
We exploit these advances to obtain 3D human-pose and an initial semantic instance segmentation from multiple view videos to bootstrap the detailed 4D understanding and modelling of complex dynamic scenes captured with multiple static or moving cameras (see Figure \ref{fig:motivation}). 
Joint 4D reconstruction allows us to understand how people move and interact, giving contextual information in general scenes.
%

%
Existing multi-task methods for scene understanding perform per frame joint reconstruction and semantic instance segmentation from a single image~\cite{KendallGC17}, showing that joint estimation can improve each task.
Other methods have fused semantic segmentation with reconstruction \cite{MustafaCVPR17} or flow estimation \cite{sevillaCVPR2016} demonstrating significant improvement in both semantic segmentation and reconstruction/scene flow. We exploit the joint estimation to understand dynamic scenes by simultaneous reconstruction, flow and segmentation estimation from multiple view video.

The first category of methods in joint estimation for dynamic scenes generate segmentation and reconstruction from multi-view \cite{Mustafa16CVPR} and monocular video \cite{Floros12,Larsen07} without any output scene flow estimate. The second category of methods segment and estimates motion in 2D \cite{sevillaCVPR2016}, or give spatio-temporal aligned segmentation \cite{chiu2013coseg,Luo2015,Djelouah16} from multiple views without retrieving the shape of the objects.
The third category of methods in 4D temporally coherent reconstruction either align meshes using correspondence information between consecutive frames \cite{Zanfir_2015_ICCV} or extract the scene flow by estimating the pairwise surface correspondence between reconstructions at successive frames \cite{Wedel2011,Basha2013}. However methods in these three categories do not exploit semantic information of the scene.
The fourth category of joint estimation methods exploit semantic information by introducing joint semantic segmentation and reconstruction for general dynamic scenes \cite{HanePAMI2016,Xie2016CVPR,Abhijit14,Ulusoy2017CVPR,MustafaCVPR17} and street scenes \cite{EngelmannGCPR16,vineet2015icra}.
However these methods give per-frame semantic segmentation and reconstruction with no motion estimate leading to unaligned geometry and pixel level incoherence in both segmentation and reconstruction for dynamic sequences.
Other methods for semantic video segmentation classify objects exploiting spatio-temporal semantic information \cite{TsaiZ016,Luo2015,chiu2013coseg} but do not perform reconstruction.
We address this gap in the literature by proposing a novel unsupervised framework for joint multi-view 4D temporally coherent reconstruction, semantic instance segmentation and flow estimation for general dynamic scenes.

Methods in the literature have exploited human-pose information to improve results in semantic segmentation \cite{Xia2017CVPR} and reconstruction \cite{MuVS3DV2017}. However existing joint methods for dynamic scenes (with multiple people) do not exploit human-pose information often detecting interacting people as a single object \cite{MustafaCVPR17}. Table \ref{t_litsurvey} shows a comparison between the tasks performed by state-of-the-art methods.
We exploit advances in 3D human-pose estimation to propose the first approach for 4D (3D in time) human-pose based scene understanding of general dynamic scenes with multiple interacting dynamic objects (people) with complex non-rigid motion. 3D human-pose estimation makes full use of multi-view information and is used as a prior to constrain the shape, segmentation and motion in space and time in the joint scene understanding estimation to improve the results.
Our contributions are:
\begin{itemize}[topsep=0pt,partopsep=0pt,itemsep=0pt,parsep=0pt] 
	\item High-level 4D scene understanding for general dynamic scenes from multi-view video.
	\item Joint instance-level segmentation, temporally coherent reconstruction and scene flow with human-pose priors.
	\item Robust 4D temporal coherence and per-pixel semantic coherence for dynamic scenes containing interactions.
	\item An extensive performance evaluation against 15 state-of-the-art methods demonstrating improved semantic segmentation, reconstruction and motion estimation.
\end{itemize}
\begin{figure*}[h]
	\begin{center}
		\includegraphics[width=0.95\textwidth]{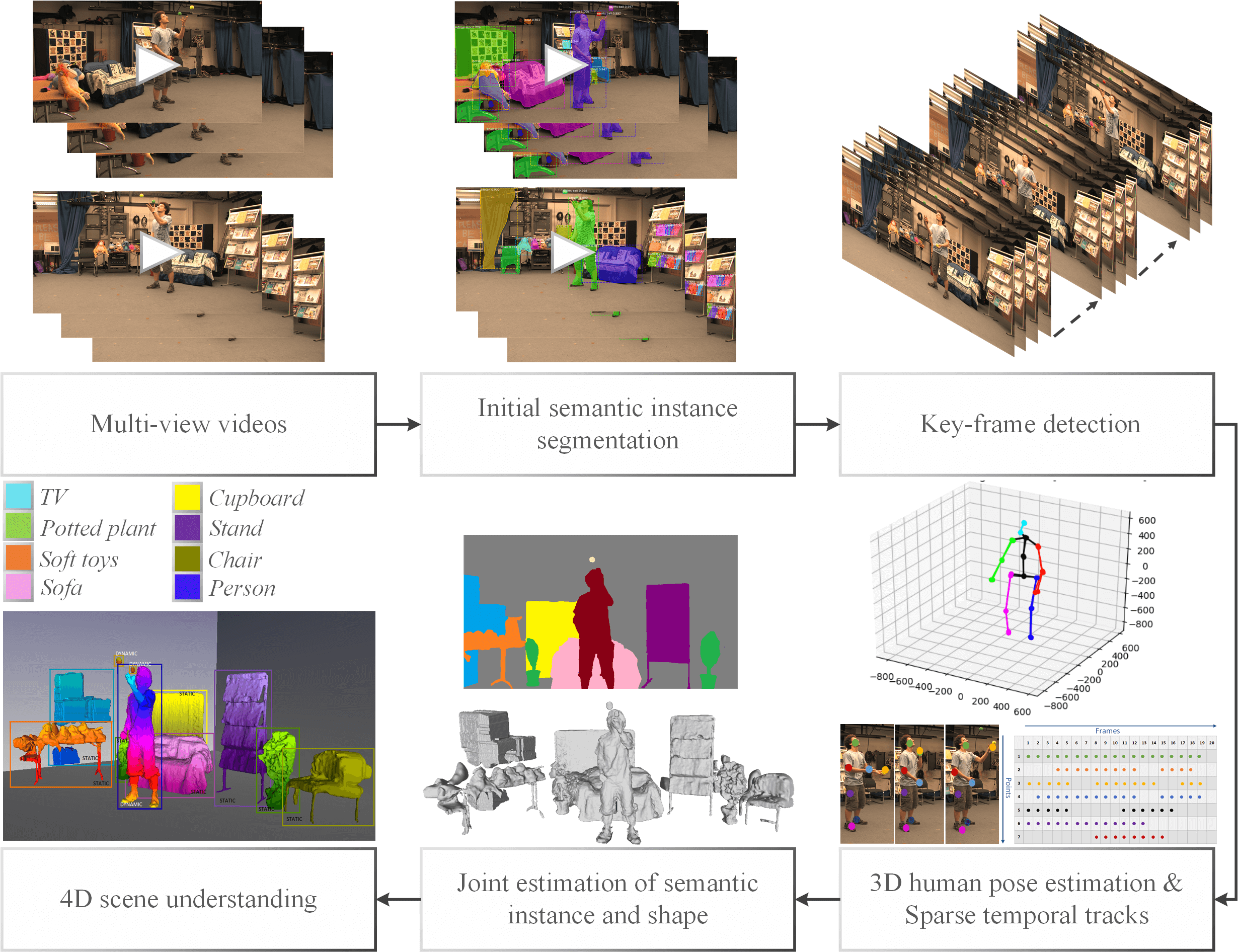}
		\caption{Unsupervised 4D scene understanding framework for dynamic scenes from multi-view video.}
		\label{fig:algorithm}
		\vspace{-0.80cm}
	\end{center}
\end{figure*}
\vspace{-0.15cm}
\section{Joint 4D dynamic scene understanding}
\label{sec:method}
%
%
This section describes our approach to joint 4D scene understanding, with different stages shown in Figure \ref{fig:algorithm}. 
The input to the joint optimisation is multi-view video, per-view initial semantic instance segmentation \cite{2017maskrcnn} and 3D human-pose estimation \cite{tome3DV2018}. 
To achieve stable long-term 4D understanding a set of unique key-frames are detected exploiting multi-view information.
Sparse temporal feature tracks are obtained per view between key-frames to initialise the joint estimation. This allows robust 4D understanding in the presence of large non-rigid motion between frames.
An initial reconstruction is obtained for each object in the scene combining the initial semantic instance segmentation with the sparse reconstruction \cite{MustafaCVPR17}.
The initial reconstruction and semantic instance segmentation is refined for each object instance through novel joint optimisation of segmentation, shape, and motion constrained by 3D human-pose (Section \ref{sec:jointrefine}).
Key-frames are used to introduce robust temporal coherence in the joint estimation across long-sequences with large non-rigid deformation.
Depth, motion and semantic instance segmentation is combined across views between frames for 4D temporally coherent reconstruction and dense per-pixel semantic coherence for final 4D understanding of scenes (Section \ref{sec:dense}).
%
\vspace{-0.15cm}
%
\begin{table}
	\setlength{\tabcolsep}{4pt}
	\scalebox{0.85}{
		\begin{tabular}{l|c|c|c|c|c|c}
			& \textbf{Semantic} & \textbf{Segment} & \textbf{Instance} & \textbf{ 3D } & \textbf{Motion} & \textbf{Pose} \\ \hline
			\cite{KendallGC17,Ulusoy2017CVPR,EngelmannGCPR16} & \checkmark & \checkmark & \checkmark & \checkmark & \textcolor{red}{$\times$} & \textcolor{red}{$\times$} \\ \hline
			\cite{sevillaCVPR2016} & \checkmark & \checkmark & \checkmark & \textcolor{red}{$\times$} & \checkmark & \textcolor{red}{$\times$} \\ \hline
			\cite{MustafaCVPR17,HanePAMI2016,Abhijit14} & \checkmark & \checkmark & \textcolor{red}{$\times$} & \checkmark & \textcolor{red}{$\times$}  & \textcolor{red}{$\times$} \\ \hline
			\cite{Xia2017CVPR} & \checkmark & \checkmark & \checkmark & \textcolor{red}{$\times$} & \textcolor{red}{$\times$} & \checkmark \\ \hline
			\cite{MuVS3DV2017} & \textcolor{red}{$\times$} & \textcolor{red}{$\times$} & \textcolor{red}{$\times$} & \checkmark & \checkmark & \checkmark \\ \hline
			\cite{Floros12} & \checkmark & \checkmark & \textcolor{red}{$\times$} & \checkmark & \checkmark & \textcolor{red}{$\times$} \\ \hline
			\cite{Larsen07,Agapito2012} & \textcolor{red}{$\times$}  & \textcolor{red}{$\times$} & \checkmark &  \checkmark &  \checkmark & \textcolor{red}{$\times$} \\ \hline
			\cite{Mustafa16CVPR} & \textcolor{red}{$\times$} & \checkmark & \textcolor{red}{$\times$} & \checkmark & \checkmark & \textcolor{red}{$\times$}  \\ \hline
			\cite{TsaiZ016,Luo2015,chiu2013coseg} & \checkmark & \checkmark & \textcolor{red}{$\times$} & \textcolor{red}{$\times$} & \checkmark & \textcolor{red}{$\times$}  \\ \hline
			\textbf{Proposed} & \checkmark & \checkmark & \checkmark & \checkmark & \checkmark & \checkmark 
	\end{tabular}}
	\caption{Comparison of tasks state-of-the-art methods are solving against the proposed method.}
	\label{t_litsurvey}
\end{table}
\subsection{Joint per-view optimisation}
\label{sec:jointrefine}
%
Existing methods for semantic segmentation do not give instance level segmentation
of the scene. Previous approach either segment the image followed by a per-segment object category classification \cite{MostajabiYS15,Gupta2014}, give deep per-pixel CNN features followed by per-pixel classification in the image \cite{FarabetCNL13,HariharanAGM15}  or predict semantic segmentation from raw pixels \cite{long_shelhamer_fcn} followed by conditional random fields \cite{Kundu2016CVPR,crfasrnn_iccv2015}.
A recent state-of-the-art method gives a good estimate of initial semantic instance segmentation masks from an image of complex sequence \cite{2017maskrcnn}. We employ this approach to predict initial semantic instance segmentation pre-trained parameters on MS-COCO\cite{LinMBHPRDZ14} and PASCAL VOC12 \cite{pascal-voc-2012} for each view.
Per-view semantic instance segmentation is combined across views with sparse reconstruction to obtain an initial reconstruction for each frame \cite{MustafaCVPR17}, this is refined through a joint scene understanding optimisation.

\begin{figure}[t]
	\begin{center}
		\includegraphics[width=0.49\textwidth]{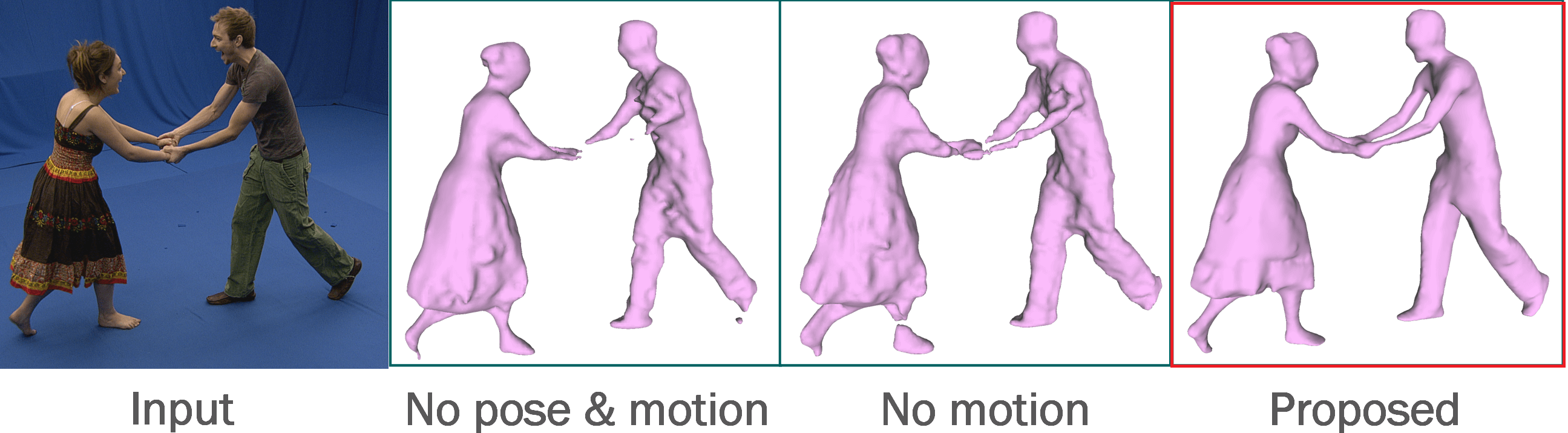}
		\caption{Comparison of reconstruction without pose and motion in the optimisation framework, proposed result is best.}
		\label{fig:pose}
		\vspace{-0.25cm}
	\end{center}
\end{figure}

%
The goal of the joint estimation is to refine initial semantic instance segmentation and reconstruction by assigning a label from a set of classes obtained from initial semantic instance segmentation $\mathscr{L} = \left \{ l_{1},...,l_{\left|\mathscr{L} \right|} \right \}$ ($\left|\mathscr{L} \right|$ is the total number of classes), a depth value from a set of depth values $\mathscr{D} = \left \{ d_{1},...,d_{\left|\mathscr{D} \right|-1}, \mathscr{U} \right \}$ (each depth value is sampled on the ray from camera and $\mathscr{U}$ is an unknown depth value to handle occlusions), and a motion flow field $\mathscr{M} = \left \{ m_{1},...,m_{\left|\mathscr{M} \right|} \right \}$ simultaneously for the region $\mathscr{R}$ of each object per view. $\left|\mathscr{M} \right|$ is the pre-defined discrete flow-fields for pixel $p = (x, y)$ in image $I$ by $m = (\delta x, \delta y)$ in time. 
Joint semantic instance segmentation, reconstruction and motion estimation is achieved by global optimisation of a cost function over unary $E_{unary}$ and pairwise $E_{pair}$ terms, defined as:
\begin{equation}\label{eq:costfunction}
E(l,d,m) = E_{unary}(l,d,m) + E_{pair}(l,d,m)
\end{equation}
\begin{equation} \nonumber
E_{unary} = \lambda _{d}E_{d}(d) + \lambda_{a}E_{a}(l) + 
\lambda _{sem}E_{sem}(l) +  \lambda _{f}E_{f}(m)
\end{equation}
\begin{equation}\nonumber
E_{pair} =  \lambda _{s}E_{s}(l,d) + \lambda _{c}E_{c}(l) +  \lambda _{r}E_{r}(l,m) + \lambda _{p}E_{p}(l,d,m) 
\end{equation}

\noindent
where, $d$ is the depth, $l$ is the class label, and $m$ is the motion at pixel $p$. 
Novel terms are introduced for flow $E_{f}$, motion regularisation $E_{r}$ and human-pose $E_{p}$ costs, explained in Section \ref{sec:motion} and \ref{sec:pose} respectively.
Results of the joint optimisation with and without pose ($E_{p}$) and motion ($E_{f}$ , $E_{r}$) information are presented in Figure \ref{fig:pose}, showing the improvement in results. Ablative analysis on individual costs in Section \ref{sec:results} show the improvement in performance with the novel introduction of motion and pose constraints in the joint optimisation.
Standard unary terms for depth ($E_{d}$), semantic ($E_{sem}$), and appearance ($E_{a}$) costs are used \cite{MustafaCVPR17}, explained in Section \ref{sec:unary}. Standard pairwise terms colour contrast ($E_{c}$) is used to assist segmentation and smoothness ($E_{s}$) cost ensures that depth vary smoothly in a neighbourhood, are explained in \textbf{\textit{Appendix A}} of the supplementary material. 

Global optimisation of Equation \ref{eq:costfunction} is performed over all terms simultaneously, using the $\alpha$-expansion algorithm by iterating through the set of labels in $\mathscr{L} \times \mathscr{D} \times \mathscr{M}$ \cite{Boykov01}. Each iteration is solved by graph-cut using the min-cut/max-flow algorithm \cite{Kolmogorov04}. Convergence is achieved in 7-8 iterations.
\vspace{-0.5cm}
\subsubsection{Spatio-temporal coherence in the optimisation}\label{sec:temporal}
\vspace{-0.15cm}
Constraints are applied on the spatial and temporal neighborhood to enforce consistency in the appearance, semantic label, 3D human pose and motion across views and time. \\
\textbf{Spatial coherence:} Multi-view spatial coherence is enforced in the optimisation such that the motion, shape, appearance, 3D pose and class labels are consistent across views using an 8-connected spatial neighbourhood $\psi_S$ for each camera view such that the set of pixel pairs $(p; q)$ belong to the same frame.\\
\textbf{Temporal coherence:} Temporal coherence is enforced in the joint optimisation by enforcing coherence across key-frames to handle large non-rigid motion and to reduce errors in sequential alignment for long sequences in the 4D scene understanding.
Sparse temporal feature correspondences are used for key-frame detection and robust initialisation of the joint optimisation. They measure the similarity between frames and unlike optical flow are robust to large motions and visual ambiguity.
To achieve robust temporal coherence in the 4D scene understanding framework for large non-rigid motion, sparse temporal feature correspondences in 3D are obtained across the sequence.

The temporal neighbourhood is defined for each frame between its respective key-frames.
Sparse temporal correspondence tracks define the temporal neighbourhood $\psi_T = \left \{ \left ( p,q \right )\mid q = p + e_{i,j} \right \}$; where $j = \left \{ t-1, t+1 \right \}$ and $e_{i,j}$ is the displacement vector from image $i$ to $j$.
\vspace{-0.25cm}
\subsubsection{Human-pose constraints $E_{p}(l,d,m)$}\label{sec:pose}
We use 3D human-pose to constrain joint optimisation and improve the flow, reconstruction and instance segmentation, in both 2D and 3D for dynamic scenes with multiple interacting people (see Figure \ref{fig:motivation}). 
3D human-pose is used as it is consistent across multiple views unlike 2D human-pose. A state-of-the-art method for 3D human-pose estimation from multiple cameras \cite{tome3DV2018} is used in the paper.
Previous work on 3D pose estimation \cite{Tome_2017_CVPR} iteratively builds a 3D model of human-pose consistent with 2D estimates of joint locations and prior knowledge of natural body pose. In \cite{tome3DV2018}, multiple cameras are used when estimating the 3D model; this then feeds back into new estimates of the 2D joint locations in each image. This approach allows us to take full advantage of 3D estimates of pose, consistent across all cameras when finding fine grained 2D correspondences between images, and leading to more lifelike, vivid human reconstructions.

Initial semantic reconstruction is updated if the 3D pose of the person lies outside the region $\mathscr{R}$ by dilating the boundary to include the missing joints. This allows for more robust and complete reconstruction and segmentation. We use a standard set of 17 joints \cite{tome3DV2018} defined as $\mathscr{B}$. A circle $\mathscr{C}_{i}$ is placed around the joint position in 2D and a sphere $\mathscr{S}_{i}$ is placed around the joint position in 3D based on the confidence map to identify the nearest neighbour vertices for every joint $b_i$.
\vspace{-0.25cm}
\begin{equation}
E_{p}(l,d,m) = \sum_{b_{i} \in \mathscr{B}} \lambda_{2d} e_{2d}(l,m) + \lambda_{3d} e_{3d}(d)
\end{equation}
\begin{equation}\nonumber
e_{2d}(l,m) =  e_{2d}^{L}(l) + e_{2d}^{S}(l) + e_{2d}^{M}(m)
\end{equation}
\begin{equation}\nonumber
e_{3d}(d) = e_{3d}^{M}(d) + e_{3d}^{S}(d) \text{, if }d_{p}\neq \mathscr{U} \text{ else } 0
\end{equation}
\textbf{3D shape term:}
This term constrains the reconstruction in 3D such that the neighbourhood points around the joints do not move far from the respective joints, and is defined as: 
\vspace{-0.25cm}
\begin{equation}\nonumber
e_{3d}^{S}(d) = \exp({ - \frac{1}{\left | \sigma_{S_{D}} \right |} \sum_{\Phi(p) \in \mathscr{S}_{i}} \left \| O \right \|_{F} ^{2} })
\vspace{-0.25cm}
\end{equation} 
where $\Phi(p)$ is the 3D projection of pixel $p$. The Frobenius norm $\left \| O \right \|_{F} = \left \| \begin{bmatrix}
\Phi(p) & b_{i} \end{bmatrix} \right \|_{F}$ is applied on the 3D points in all directions to obtain the ‘net’ motion at each pixel within $\mathscr{S}_{i}$ and $\sigma_{S_{D}} = \left \langle\frac{\left \| O \right \|_{F}^{2}}{\vartheta_{\Phi(p),b_i}}\right\rangle$.\\
%
\textbf{3D motion term:}
This enforces as rigid as possible \cite{Sorkine2007} constraint on 3D points in the neighbourhood of each joint $b_{i}$ in space and time. An optimal rotation matrix $R_{i}$ is estimated for each $b_{i}$ by minimising the energy defined as:
\vspace{-0.25cm}
\begin{multline}\nonumber
e_{3d}^{M}(d) = \sum_{\Phi(p) \in \mathscr{S}_{i}} \left \| \left ( b_{i}^{t+1} - \Phi(p)^{t+1} \right ) - R_{i} \left ( b_{i}^{t} - \Phi(p)^{t} \right ) \right \|_{2}^{2} \\
+ \lambda _{3d}^{p} \left \| p -  e_{3d}^{M} \right \|_{2}^{2}
\end{multline} 
%
\textbf{2D term:}
3D poses are back-projected in each view to constrain per view appearance ($e_{2d}^{L}$), semantic segmentation ($e_{2d}^{S}$) and motion estimation ($e_{2d}^{M}$) in 2D. If $p \in \mathscr{C}_{i}$,
\vspace{-0.25cm}
\begin{multline}\nonumber
e_{2d}^{L}(l) = \exp\left ( -\sum_{p \in \psi_S}  \sum_{p \in \psi_T} \frac{\left \| I(\Pi(b_i)) - I(p) \right \|^{2}}{\left | \sigma_{S_{L}} \right |}\right ) \\
e_{2d}^{S}(l) = \exp \left ( -\sum_{p \in \psi_S}  \sum_{p \in \psi_T} \frac{\left \| \Pi(b_i) - p \right \|^{2}}{\left | \sigma_{S_{S}} \right |}\right ) \\
e_{2d}^{M}(m) = \exp \left (-\sum_{p \in \psi_S} \sum_{k \in \psi_T} \frac{\left \| \vartheta_{p,\Pi (b^{k}_{i})} - \vartheta_{p+m_{p},\Pi (b^{k+1}_{i})} \right \|^{2}}{\left | \sigma_{S_{M}} \right |} \right )
\end{multline}
where, $\Pi$ is the back-projection of 3D poses to 2D, $N_{pose}$ is the number of nearest neighbours, $\sigma_{S_{L}} = \left \langle\frac{\left \| \Pi(b_i) - q \right \|^{2}}{\vartheta_{\Pi(b_i),q}}\right\rangle$ and, $\sigma_{S_{S}}$ and $\sigma_{S_{M}}$ is defined similarly. $e_{2d}^{L}(l)$ and $e_{2d}^{S}(l)$ ensures that the pixels around projected 3D pose $\Pi (b_{i})$ have the same semantic label and appearance across views ($\psi_S$) and time ($\psi_T$) thereby ensuring spatio-temporal appearance and semantic consistency respectively.
\vspace{-0.15cm}
\subsubsection{Motion constraints- $E_{f}(m) \text{ and } E_{r}(l,m)$} 
\label{sec:motion}
%
%
\textbf{Flow term:} This term is obtained by integrating the sum of three penalisers over the reference image domain inspired from \cite{Tao2012}, defined as:\\
$E_{f}({p,m_p}) = e_{F}^{T}({p,m_p}) + e_{F}^{V}({p,m_p}) + e_{F}^{S}({p,m_p})$ \\ where, $e_{F}^{T}({p,m_p}) =\sum_{i = 1}^{N_{v}} \left \| (I_{i}(p,t) - I_{i}(p+ m_p, t+1)) \right \| ^{2}$ penalises deviation from the brightness constancy assumption in a temporal neighbourhood for the same view; $e_{F}^{V}({p,m_p}) =\sum_{t \in \psi_T} \sum_{i = 2}^{N_{v}} \left \| (I_{1}(p,t) - \\ I_{i}(p+ m_p,t)) \right \| ^{2}$ penalises deviation in appearance from the brightness constancy assumption between the reference view and other views at other time instants; and $ e_{F}^{S}({p,m_p}) = 0 \text{ if } p \in N \text{ otherwise } \infty $ which forces the flow to be close to nearby sparse temporal correspondences. $I_{i}(p,t)$ is the intensity at point $p$ at time $t$ in camera $i$.
The flow vector $m$ is located within a window from a sparse constraint at $p$ and it forces the flow to approximate the sparse 2D temporal correspondences.
\\
\noindent
\textbf{Motion regularisation term:} This penalises the absolute difference of the flow field to enforce motion smoothness and handle occlusions in areas with low confidence \cite{Tao2012}.\\
$E_{r}({l,m}) = \sum_{p,q \in N_{p}} \left \| \Delta m \right \|^{2} \lambda_{r}^{L} e_{r}^{L} (p, q, m_{p}, m_{q}, l_{p}, l_{q}) + \lambda_{r}^{A} e_{r}^{A} (p, q, m_{p}, m_{q}, l_{p}, l_{q})$ \\
where  $\Delta m = m_{p} - m_{q}$ and; \\ 
$e_{r}^{X} = \underset{l_{p} = l_{q}}{\forall} \text{ }\underset{q \in N_{p}}{\text{mean}} \text{ } E_{X}({q,m_q}) - \underset{q \in N_{p}}{\min} E_{X}({q,m_q})$ else $0$.
We compute $e_{R}^{L}$ (semantic regularisation) and $e_{R}^{A}$ (appearance regularisation) as the minimum subtracted from the mean energy within the search window $N_{p}$ for each pixel $p$.
%
%
%
%
%
\vspace{-0.3cm}
\subsubsection{Long-term temporal coherence}\label{sec:keyframe}
\vspace{-0.15cm}
\textbf{Sparse temporal correspondences:} The sparse 3D points projected in all views are matched between frames $N_{f}^{i}$ and key-frames across the sequence using nearest neighbour matching \cite{Lowe04} followed by a symmetry test which employs forward and backward match consistency by performing two-way matching to remove the inconsistent correspondences. This gives sparse temporal feature correspondence tracks per frame for each object:
$F^{c}_{i} = \{ {f^{c}_{1}, f^{c}_{2}, . . . , f^{c}_{R_{i}^{c}}}\}$, where $c={1 \text{ to } N_{v}}$. $R_{i}^{c}$ are the 3D points visible at each frame $i$.
Exhaustive matching is done, such that each frame is matched to every other frame to handle appearance, reappearance and disappearance of points between frames.

\begin{figure}
	\begin{center}
		\includegraphics[width=0.49\textwidth]{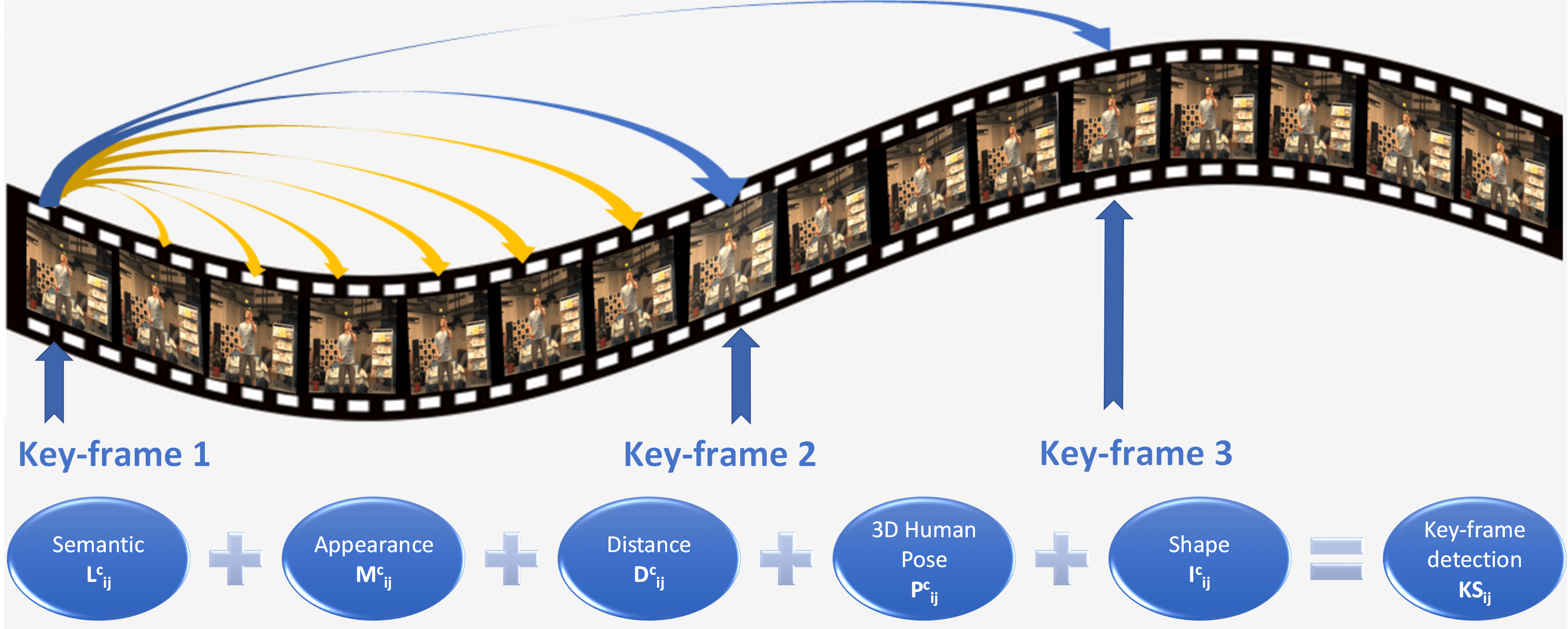}
		\caption{An illustration of key-frame detection and matching across a short sequence for stable long-term temporal coherence.}
		\label{fig:keyframe}
		\vspace{-0.25cm}
	\end{center}
\end{figure}

\noindent
\textbf{Key-frame detection:} Previous work \cite{Newcombe2015DynamicFusionRA,Mustafa3DV17} showed that sparse key-frames allow robust long-term correspondence for 4D reconstruction. In this work we introduce the additional use of pose in the detection and sparse temporal feature correspondence across key-frames to prevent the accumulation of errors in long sequences.
4D scene alignment between key-frames is explained in Section \ref{sec:dense}. \\
\textit{Key-frame similarity metric} is defined as: \\
\vspace{-0.3cm}
\begin{equation}\label{eq:keyframe}
KS_{i,j} = 1 - \frac{1}{5N_{v}}\sum_{c=1}^{N_{v}} ( M_{i,j}^{c} + L_{i,j}^{c} + D_{i,j}^{c} + P_{i,j}^{c} + I_{i,j}^{c})
\vspace{-0.25cm}
\end{equation}
Key-frame detection exploits sparse correspondence ($M_{i,j}^{c}$), pose ($P_{i,j}^{c}$), shape ($I_{i,j}^{c}$), semantic ($I_{i,j}^{c}$) and distance ($D_{i,j}^{c}$) information across views $N_{v}$ between frame $i$ and $j$ for each object in view $c$, to improve the long-term temporal coherence of the proposed method, using similar frames across the sequence, illustrated in Figure \ref{fig:keyframe}.
All frames with similarity $>0.75$ in a sequence are selected as key-frames defined as $K = \{ {k^{1}, k^{2}, . . . , k^{N_{k}}}\}$ where $N_{k}$ is the number of key-frames and $N_{f}^{i}$ is the number of frames between $K_{i}$ and $K_{i+1}$.
All the metrics used in \ref{eq:keyframe} and an ablation study for key-frame detection is given in detail in \textbf{\textit{Appendix B}} of supplementary material.

Features at view $c$ frame $i$, $F^{c}_{i}$ are matched to features at view $c$ to frames $j = \{ {i+1, . . . , N_{f}^{i}}\}$ to give correspondences for all the frames $N_{f}^{i}$ with key-frame $K_{i}$. The corresponding joint locations from the 3D pose are back-projected in each view and added to sparse temporal tracks in between key-frames. Any new point-tracks are added to the list of point tracks for key-frame $K_{i}$.
%
%
%
%
%
\vspace{-0.15cm}
\subsubsection{Unary terms - $E_{unary}(l,d,m)$} 
\label{sec:unary}
\vspace{-0.15cm}
\textbf{Depth term:}
This gives a measure of photo-consistency between views $E_{d}(d) = \sum_{p\in \psi_S} e_{d}(p, d_{p})$, defined as:\\
\vspace{-0.25cm}
\begin{equation*} \label{eq:matching1}
e_{d}(p, d_{p}) = 
\begin{cases}
M(p, q)  = \sum_{i \in \mathscr{O}_{k}}m(p,q) ,& \text{if } d_{p}\neq \mathscr{U}\\
M_{\mathscr{U}}, & \text{if } d_{p} = \mathscr{U}
\end{cases}
\vspace{-0.25cm}
\end{equation*}
where $M_{\mathscr{U}}$ is the fixed cost of labelling pixel unknown and $q$ denotes the projection of the hypothesised point $P$ ($3D$ point along the optical ray passing through pixel $p$ located at a distance $d_{p}$ from the camera) in an auxiliary camera. $\mathscr{O}_{k}$ is the set of the $k$ most photo-consistent pairs with reference camera and $m(p,q)$ is inspired from \cite{Mustafa16CVPR}.\\
%
\textbf{Appearance term:}
This term is computed using the negative log likelihood \cite{Kolmogorov04} of the colour models (GMMs with 10 components) learned from the initial semantic mask in the temporal neighbourhood $\psi_T$ and the foreground markers obtained from the sparse 3D features for the dynamic objects. It is defined as: \\
$E_{a}(l) = \sum_{p \in \psi_T} \sum_{p\in \psi_S} -\log P(I_{p}\rvert l_{p})$\\
where $P(I_{p}\rvert l_{p} = l_i)$ denotes the probability of pixel $p$ belonging to layer $l_i$.\\
%
\textbf{Semantic term:}
This term is based on the probability of the class labels at each pixel based on \cite{ChenPK0Y16}, defined as:\\ $E_{sem}(l) = \sum_{p \in \psi_T} \sum_{p\in \psi_S} -\log P_{sem}(I_{p}\rvert l_{p})$\\
where $P_{sem}(I_{p}\rvert l_{p} = l_i)$ denotes the probability of pixel $p$ being in layer $l_i$ in the reference image obtained from initial semantic instance segmentation \cite{2017maskrcnn}.
%
%
\begin{figure}
	\begin{center}
		\includegraphics[width=0.49\textwidth]{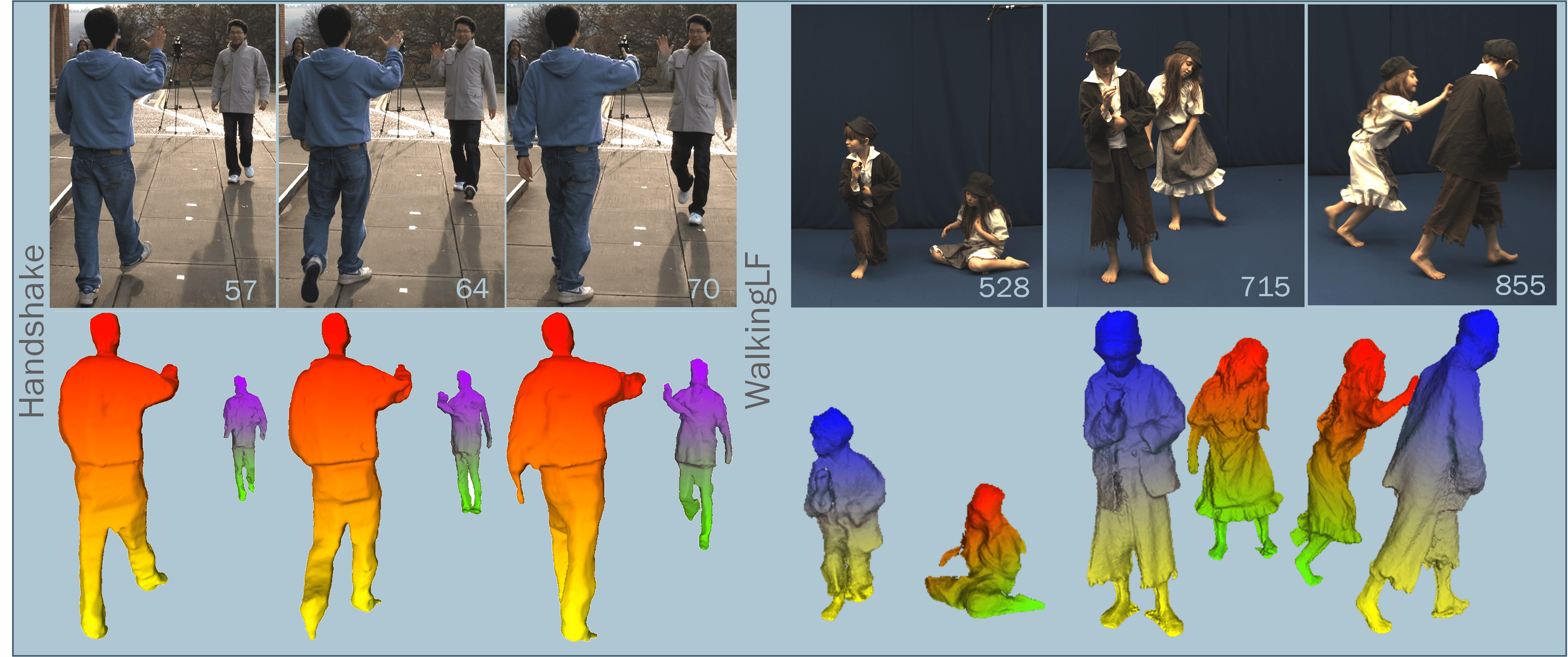}
		\caption{Example of 4D scene reconstruction for two datasets}
		\label{fig:4D}
		\vspace{-0.05cm}
	\end{center}
\end{figure}
\begin{table}
	\centering
	\setlength{\tabcolsep}{1pt}
	\scalebox{0.9}{
		\begin{tabular}{l|c|c|c|c|c|c}
			\textbf{Datasets} &  \textbf{Resolution} & \textbf{$N_{v}$} & \textbf{Baseline} &  \textbf{L} & \textbf{KF} &\textbf{Tracks} \\ \hline
			Handshake\cite{Kim2012} & $1920 \times 1080$ & 8(all S) & $15\degree$-$30\degree$ & 125 & 15 & 1945 \\
			Meetup\cite{Guillemaut2010} & $1920 \times 1080$ & 16(all S) & $25\degree$-$35\degree$ & 100 & 9 & 1341 \\
			Juggler2\cite{UnstructuredVBR10} & $960 \times 544$ & 6(all M) & $15\degree$-$45\degree$ & 300 & 16 & 1278 \\
			Handstand\cite{Vlasic2008} & $1600 \times 1200$ & 8(all S) & $25\degree$-$45\degree$ & 174 & 12 & 1056\\
			Rachel\cite{cvssp3d} & $3840 \times 2160$ & 16(all S) & $20\degree$-$30\degree$ & 270 & 15 & 1978\\
			Juggler1\cite{cvssp3d} & $1920 \times 1080$ & 8(2 M) & $15\degree$-$30\degree$ & 253 & 17 & 2083 \\
			Dance\cite{4DInria} & $780 \times 582$ & 8(all S) & $35\degree$-$45\degree$ & 60 & 7 & 732 \\
			Magician\cite{UnstructuredVBR10}& $960 \times 544$ & 6(all M) & $15\degree$-$45\degree$ & 300 & 10 & 1312 \\
			Human3.6\cite{h36m_pami} & $1000 \times 1000$ & 4(all S) & $25\degree$-$30\degree$ & 250 & 14 & 994 \\
			MagicianLF\cite{Mustafa3DV17} & $2048 \times 2048$ & 25(all S) & $5\degree$-$8\degree$ & 350 & 5 & 1312 \\
			WalkLF\cite{Mustafa3DV17} & $2048 \times 2048$ & 20(all S) & $5\degree$-$8\degree$ & 221 & 7 & 1934 \\
	\end{tabular}}
	\caption{Properties of all datasets: $N_{v}$ is the number of views, L is the sequence length, KF gives number of key-frames, and Tracks gives the number of sparse temporal correspondence tracks averaged over the entire sequence for each object (S stands for static cameras and M for moving cameras).}
	\vspace{0.2cm}
	\label{t_dataset}
\end{table}
%
%
%
\section{4D scene understanding}
\label{sec:dense}
\vspace{-0.15cm}
The final 4D scene model fuses the semantic instance segmentation, depth information and dense flow across views and in time between frames ($N_{f}^{i}$) and key-frames ($K_{i}$). The initial instance segmentation, human pose and motion information for each object is combined to obtain final instance segmentation of the scene.
The depth information is combined across views using Poisson surface reconstruction \cite{Kazhdan2006} to obtain a mesh for each object in the scene.
4D temporally coherent meshes are obtained by combining the most consistent motion information from all views for each 3D point. This is combined with spatial semantic instance information to give per-pixel semantic and temporal coherence. Appearing, disappearing, and reappearing regions are handled by using the sparse temporal tracks and their respective motion estimate. 
The dense flow and semantic instance segmentation together with 3D models of each object in the scene gives the final 4D understanding of the scenes. Examples are shown in Figure \ref{fig:motivation} and \ref{fig:4D} on two datasets, where objects are coloured in one key-frame and colours are propagated reliably between frames and key-frames across the sequence for robust 4D scene modelling.
\vspace{-0.5cm}
\section{Results and evaluation}
\label{sec:results}
\vspace{-0.15cm}
Joint semantic instance segmentation, reconstruction and flow estimation (section \ref{sec:method}) is evaluated quantitatively and qualitatively against $15$ state-of-the-art methods on a variety of publically available multi-view indoor and outdoor dynamic scene datasets, detailed in Table \ref{t_dataset}. More results are provided in supplementary material \textbf{\textit{Appendix C}}.

Algorithm parameters listed in Table \ref{t_parameters} are the same for all outdoor datasets, and for indoor datasets parameters depend on the number of cameras ($N_{v}$). Pairwise costs are constant $\lambda_{p} = 0.9$, $\lambda_{c}=\lambda_{s}=\lambda_{r}=0.5$ for all datasets.

%
\subsection{Reconstruction evaluation}
\label{sec:reconstresults}
\begin{figure}
	\begin{center}
		\includegraphics[width=0.49\textwidth]{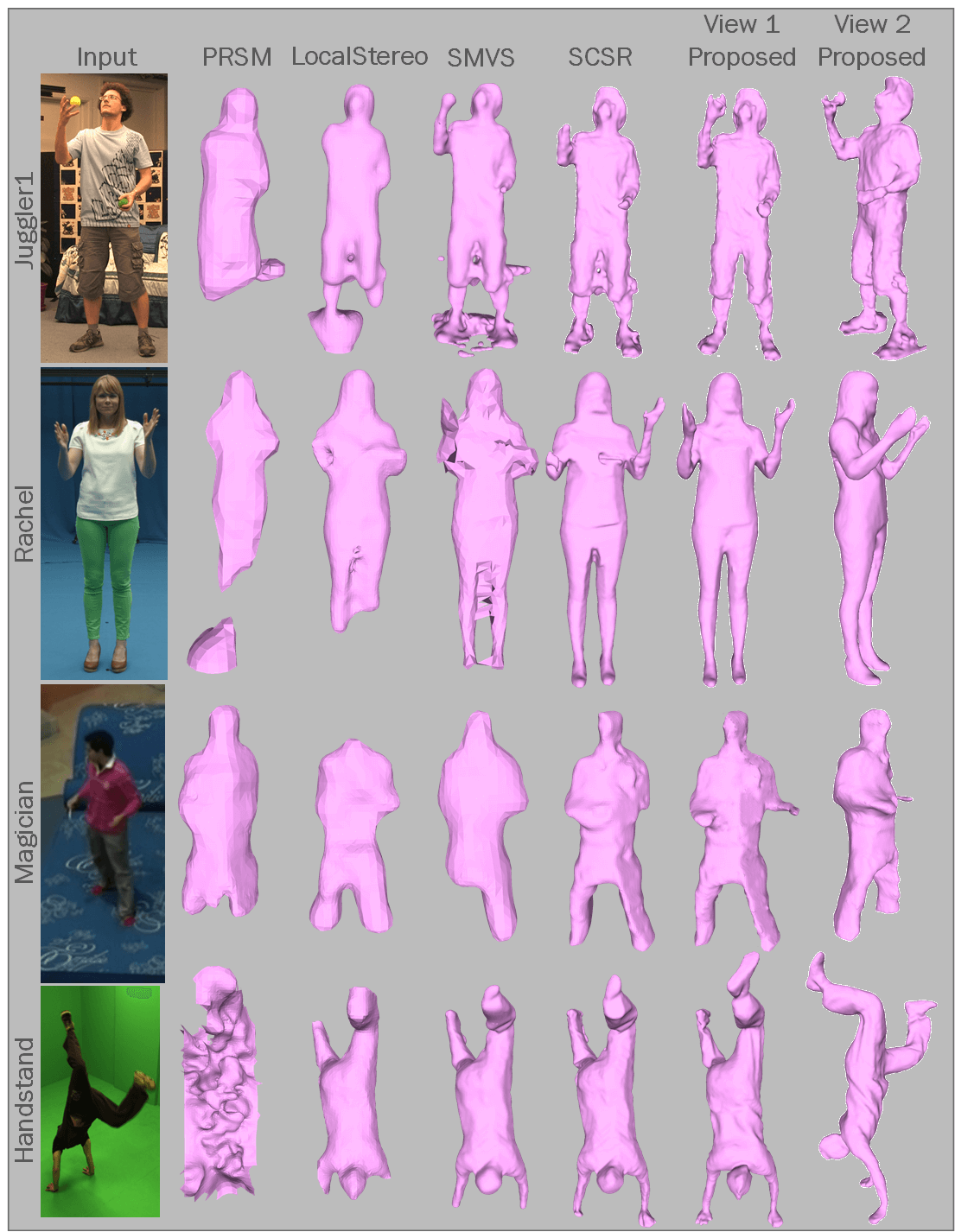}
		\caption{Reconstruction evaluation against existing methods. Two different views of 3D model are shown for proposed method.}
		\label{fig:reconstruct}
		\vspace{-0.15cm}
	\end{center}
\end{figure}

\vspace{-0.15cm}
The proposed approach is compared against state-of-the-art approaches for semantic co-segmentation and reconstruction (SCSR) \cite{MustafaCVPR17}, piecewise scene flow (PRSM) \cite{Vogel2015IJCV}, multi-view stereo (SMVS) \cite{langguth-2016-smvs}, and deep learning based stereo approaches (LocalStereo) \cite{Taniai18}. Qualitative comparison with 2 views of proposed method are shown in Figure \ref{fig:reconstruct}.
Pre-trained parameters were used for LocalStereo and per-view depth maps were fused using Poisson reconstruction. The quality of surface obtained using proposed method is improved compared to state-of-the-art methods. In contrast to previous approaches, limbs of people are reliably reconstructed because of the exploitation of human-pose and temporal information (motion) in the joint optimisation.

For quantitative comparison to state-of-the-art methods, we project the reconstruction onto different views and compute the projection errors shown in Table \ref{reconstructCompare}. A significant improvement is obtained in projected surface completeness with the proposed approach.
%
%
\begin{table}[t]
	\setlength{\tabcolsep}{2.5pt}
	\scalebox{0.85}{
		\begin{tabular}{l|c|c|c|c|c|c|c|c}
			& \multicolumn{1}{c|}{$\lambda_{d}$} & \multicolumn{1}{c|}{$\lambda_{a}$} & \multicolumn{1}{c|}{$\lambda_{sem}$} & $\lambda_{f}$ & $\lambda_{s}^{t}/\lambda_{s}^{s}$ & $\lambda_{ca}/\lambda_{cl}$ & $\lambda_{r}^{L}/\lambda_{r}^{C}$ & $\lambda_{2d}/\lambda_{3d}$ \\ \hline
			\multicolumn{1}{l|}{Outdoor} & 1.2 & 0.5 & 0.5 & 0.4 & 1.0 & 5.0 & 0.6 & 7.5 \\ 
			\multicolumn{1}{l|}{I,$N_{v}<6 $} & 1.0 & 0.7 & 0.5 & 0.6 & 0.4 & 5.0  & 0.4  & 7.5 \\ 
			\multicolumn{1}{l|}{I,$ 6 \leq N_{v} <20$} & 1.0 & 0.7 & 0.2 & 0.4 & 0.4 & 5.0 & 0.4 & 5.0 \\ 
			\multicolumn{1}{l|}{I,$N_{v}\geq 20$} & 1.0 & 1.0 & 0.5 & 0.5 & 0.2 & 5.0  & 0.4 & 5.0 \\ 
	\end{tabular}}
	\caption{Parameters for all datasets. I is Indoor}
	\vspace{0.1cm}
	\label{t_parameters}
\end{table}
%
%
\begin{table*}[]
	\centering
	\setlength{\tabcolsep}{4.5pt}
	\scalebox{0.85}{
		\begin{tabular}{l|c|c|c|c|c|c|c|c|c|c|c}
			\textbf{Methods} & \textbf{Handshake} & \textbf{Handstand} & \textbf{Rachel} & \textbf{Juggler1} & \textbf{Juggler2} & \textbf{Magician} & \textbf{Dance} & \textbf{Meetup} & \textbf{Human3.6} & \textbf{MagicianLF} & \textbf{WalkLF} \\ \hline
			PRSM \cite{Vogel2015IJCV} & 1.56 & 1.79 & 1.51 & 1.57 & 1.68 & 1.72 & 1.79 & 1.98 & 2.01 & 1.59 & 1.41 \\
			LS \cite{Taniai18} & 1.24 & 1.38 & 1.15 & 1.21 & 1.18 & 1.33 & 1.46 & 1.47 & 1.64 & 1.20 & 1.23 \\
			SMVS \cite{langguth-2016-smvs} & 0.84 & 0.97 & 0.73 & 0.75 & 0.85 & 0.92 & 0.85 & 0.96 & 1.19 & 0.94 & 0.88 \\
			SCSR \cite{MustafaCVPR17} & 0.70 & 0.84 & 0.67 & 0.69 & 0.73 & 0.78 & 0.77 & 0.87 & 0.92 & 0.77 & 0.71 \\
			$P_{PS}$ & 0.73 & 0.87 & 0.65 & 0.70 & 0.71 & 0.75 & 0.74 & 0.88 & 0.90 & 0.78 & 0.70 \\
			$P_{PM}$ & 0.71 & 0.85 & 0.64 & 0.68 & 0.69 & 0.73 & 0.72 & 0.85 & 0.87 & 0.75 & 0.68 \\
			$P_{P}$ & 0.57 & 0.71 & 0.56 & 0.59 & 0.61 & 0.64 & 0.62 & 0.75 & 0.77 & 0.67 & 0.63 \\
			$P_{S}$ & 0.59 & 0.69 & 0.59 & 0.57 & 0.63 & 0.66 & 0.60 & 0.73 & 0.76 & 0.65 & 0.60 \\
			$P_{M}$ & 0.55 & 0.68 & 0.55 & 0.54 & 0.59 & 0.61 & 0.59 & 0.74 & 0.73 & 0.62 & 0.59 \\			
			Proposed & \textbf{0.46} & \textbf{0.55} & \textbf{0.47} & \textbf{0.49} & \textbf{0.51} & \textbf{0.53} & \textbf{0.55} & \textbf{0.57} & \textbf{0.60} & \textbf{0.49} & \textbf{0.44}\\
	\end{tabular}}
	\caption{Reconstruction evaluation: Projection error across views against state-of-the-art methods, LS is LocalStereo. $P_{P} = E- E_{p},P_{M} = E- E_{f} - E_{r},P_{PM} = E- E_{f} - E_{r} - E_{p},P_{S} = E- E_{sem}$ and $P_{PS} = E- E_{sem} - E_{p}$, where $E$ is defined in Equation \ref{eq:costfunction}.}
	\vspace{-0.35cm}
	\label{reconstructCompare}
\end{table*}
%
%
%
\begin{table*}[]
	\centering
	\setlength{\tabcolsep}{4pt}
	\scalebox{0.85}{
		\begin{tabular}{l|c|c|c|c|c|c|c|c|c|c|c}
			\textbf{Methods}  & \textbf{Handshake} & \textbf{Handstand} & \textbf{Rachel} & \textbf{Juggler1} & \textbf{Juggler2} & \textbf{Magician} & \textbf{Dance} & \textbf{Meetup} & \textbf{Human3.6} & \textbf{MagicianLF} & \textbf{WalkLF} \\ \hline
			CRFRNN \cite{crfasrnn_iccv2015} & 62.7 & 55.8 & 61.6 & 40.5 & 68.7 & 52.4 & 49.3 & 41.1 & 42.9 & 60.8 & 63.6 \\
			Segnet \cite{badrinarayanan2015segnet} & 47.9 & 51.1 & 55.2 & 45.1 & 61.9 & 55.3 & 53.9 & 43.9 & 49.4 & 59.3 & 65.9 \\
			JSR \cite{Guillemaut2010} & 67.8 & 58.7 & 58.4 & 56.2 & 66.0 & 61.3 & 57.9 & 50.2 & 53.4 & 62.3 & 68.9 \\
			SCV \cite{TsaiZ016} & 56.4 & 52.6 & 48.8 & 49.5 & 59.1 & 59.2 & 56.7 & 42.0 & 49.1 & 58.2 & 65.7 \\
			Dv3+ \cite{deeplabv3} & 63.8 & 58.9 & 64.0 & 48.8 & 69.7 & 58.9 & 57.6 & 48.4 & 54.8 & 69.6 & 69.1 \\
			MRCNN \cite{2017maskrcnn} & 65.2 & 59.6 & 67.4 & 50.3 & 70.5 & 60.5 & 58.7 & 47.2 & 53.4 & 69.5 & 70.2 \\
			PSP \cite{zhao2017pspnet}  & 74.7 & 64.5 & 75.5 & 67.9 & 81.2 & 73.4 & 71.5 & 62.6 & 65.3 & 74.6 & 82.5 \\
			SCSR \cite{MustafaCVPR17} & 81.8 & 75.2 & 78.4 & 81.4 & 89.3 & 88.2 & 85.1 & 78.9 & 70.4 & 82.2 & 86.7 \\
			$P_{PM}$ & 85.7 & 75.9 & 78.6 & 81.8 & 89.6 & 88.5 & 85.5 & 79.2 & 70.6 & 82.9 & 87.5 \\
			$P_{P}$ & 86.3 & 77.4 & 80.7 & 82.6 & 90.1 & 89.1 & 87.6 & 80.8 & 76.3 & 86.1 & 89.3 \\
			$P_{M}$ & 87.6 & 79.1 & 81.7 & 83.5 & 90.5 & 89.6 & 86.4 & 81.9 & 75.4 & 85.2 & 88.1 \\
			Proposed & \textbf{89.6} & \textbf{83.3} & \textbf{85.8} & \textbf{88.2} & \textbf{91.1} & \textbf{90.9} & \textbf{88.5} & \textbf{84.7} & \textbf{81.1} & \textbf{89.4} & \textbf{91.8}
	\end{tabular}}
	\caption{Segmentation comparison against state-of-the-art methods using the \textit{Intersection-over-Union} metric.}
	\label{segTable}
	\vspace{-0.5cm}
\end{table*}
\begin{figure*}[t]
	\begin{center}
		\includegraphics[width=0.98\textwidth]{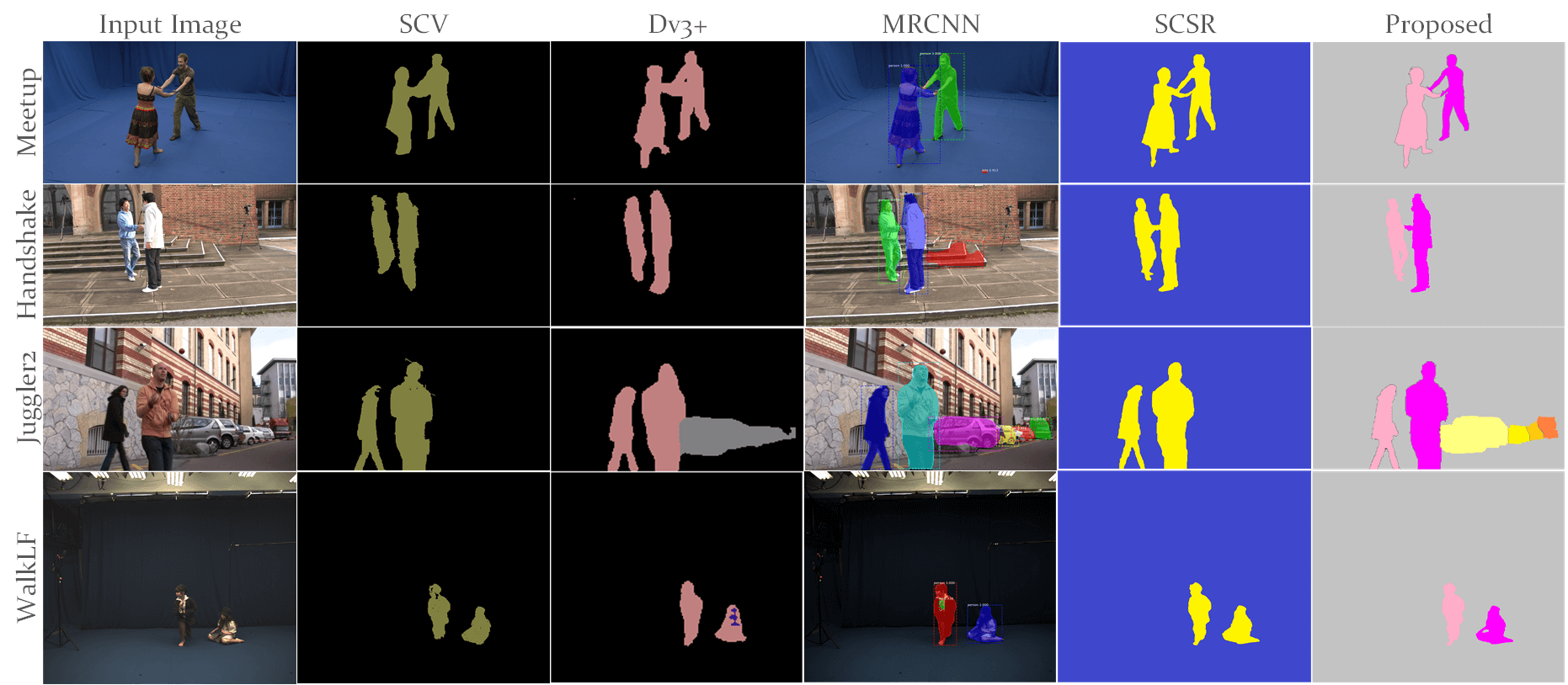}
		\caption{Semantic segmentation comparison against state-of-the-art methods. In the proposed method shades of pink depicts instances of humans and shades of yellow depict instances of cars.}
		\label{fig:segment}
		\vspace{-0.70cm}
	\end{center}
\end{figure*}
%
\begin{table*}
	\centering
	\setlength{\tabcolsep}{4pt}
	\scalebox{0.85}{
		\begin{tabular}{l|c|c|c|c|c|c|c|c|c|c|c}
			\textbf{Methods}  & \textbf{Handshake} & \textbf{Handstand} & \textbf{Rachel} & \textbf{Juggler1} & \textbf{Juggler2} & \textbf{Magician} & \textbf{Dance} & \textbf{Meetup} & \textbf{Human3.6} & \textbf{MagicianLF} & \textbf{WalkLF} \\ \hline
			PRSM \cite{XRK2017} & 1.80 & 2.15 & 1.54 & 1.65 & 1.79 & 1.96 & 1.87 & 2.11 & 2.34 & 1.87 & 1.52 \\
			Deepflow \cite{deepflow} & 1.15 & 1.48 & 1.01 & 1.08 & 1.16 & 1.27 & 1.21 & 1.37 & 1.52 & 1.05 & 0.81 \\
			DCFlow \cite{Vogel2015IJCV} & 0.90 & 1.17 & 0.97 & 0.87 & 0.93 & 1.03 & 0.96 & 1.12 & 1.21 & 0.83 & 0.79 \\
			4DMatch \cite{Mustafa16ECCV} & 0.79 & 0.98 & 0.75 & 0.69 & 0.87 & 0.81 & 0.77 & 0.87 & 0.94 & 0.80 & 0.77 \\
			$P_{PS}$& 0.75 & 1.01 & 0.85 & 0.78 & 0.91 & 0.93 & 0.86 & 0.99 & 1.07 & 0.81 & 0.78 \\
			$P_{P}$ & 0.71 & 0.93 & 0.80 & 0.73 & 0.84 & 0.87 & 0.78 & 0.92 & 0.99 & 0.76 & 0.73 \\
			$P_{S}$ & 0.64 & 0.77 & 0.63 & 0.61 & 0.65 & 0.72 & 0.65 & 0.76 & 0.81 & 0.64 & 0.61 \\
			Proposed & \textbf{0.51} & \textbf{0.61} & \textbf{0.48} & \textbf{0.49} & \textbf{0.52} & \textbf{0.58} & \textbf{0.55} & \textbf{0.63} & \textbf{0.68} & \textbf{0.53} & \textbf{0.44}
	\end{tabular}}
	\caption{Silhouette overlap error for multi-view datasets for evaluation of long-term temporal coherence, where .}
	\label{soe}
	\vspace{-0.65cm}
\end{table*}
%
\subsection{Segmentation evaluation}
\label{sec:segresults}
\vspace{-0.15cm}
Our approach is evaluated against a variety of state-of-the-art multi-view (SCV \cite{TsaiZ016}, SCSR \cite{MustafaCVPR17}, and JSR \cite{Guillemaut2010}) and single-view (Dv3+ \cite{deeplabv3}, MRCNN \cite{2017maskrcnn}, PSP \cite{zhao2017pspnet}, CRF RNN \cite{crfasrnn_iccv2015}, and Segnet \cite{badrinarayanan2015segnet}) segmentation methods, shown in Figure \ref{fig:segment}. For fair evaluation against single-view semantic segmentation methods, multi-view consistency is applied for segmentation estimated from each view to obtain multi-view consistent semantic segmentation using dense multi-view correspondence. Colour in the results is kept from the original papers. Only MRCNN and the proposed approach gives instance segmentation.
%

Quantitative evaluation against state-of-the-art methods is measured by \textit{Intersection-over-Union} with ground-truth, shown in Table \ref{segTable}. Ground-truth is available on-line for most of the datasets and obtained by manual labelling for other datasets. Pre-trained parameters were used for semantic segmentation methods. The semantic instance segmentation results from the joint optimisation are significantly better compared to the state-of-the-art methods ($\approx 20 - 40 \%$).
%
\begin{figure}
	\begin{center}
		\includegraphics[width=0.46\textwidth]{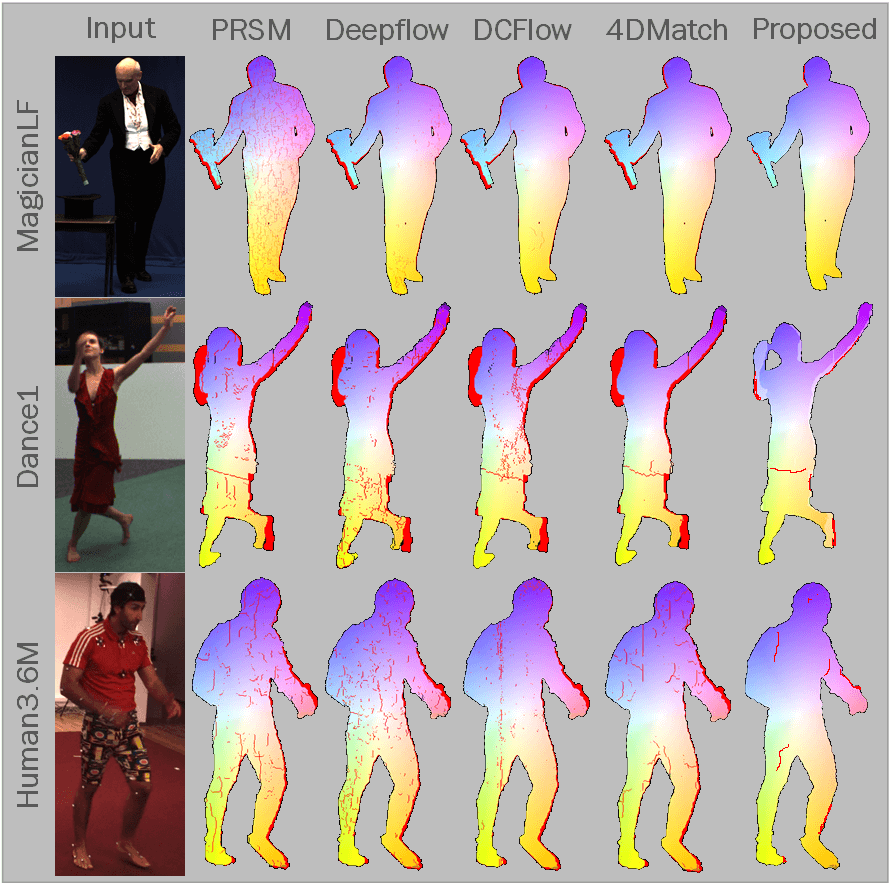}
		\caption{Temporal coherence evaluation against existing methods.}
		\label{fig:flow}
		\vspace{-0.25cm}
	\end{center}
\end{figure}
\begin{figure}
	\begin{center}
		\includegraphics[width=0.49\textwidth]{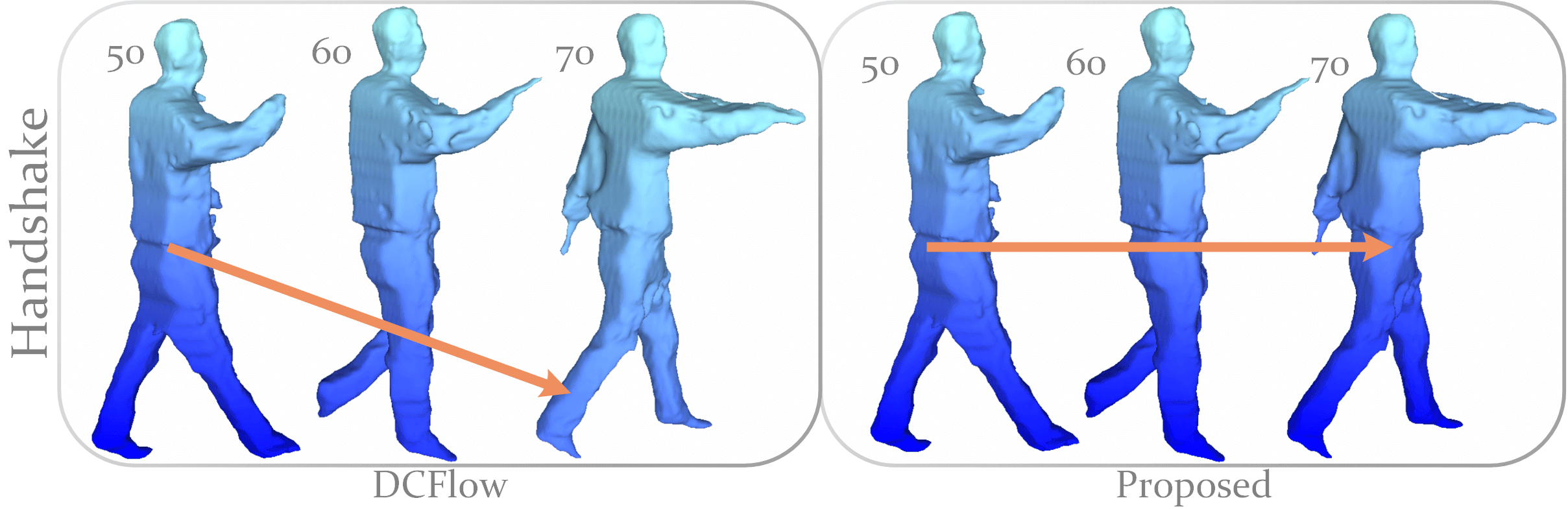}
		\vspace{-0.50cm}
		\caption{4D alignment evaluation against DCFlow \cite{XRK2017}.}
		\label{fig:4Deval}
		\vspace{-0.15cm}
	\end{center}
\end{figure}
\vspace{-0.5cm}
%
\subsection{Motion evaluation}
\label{sec:motionresults}
\vspace{-0.15cm}
Flow from the joint estimation is evaluated against state-of-the-art methods: (a) Dense flow algorithms DCflow \cite{XRK2017} and Deepflow \cite{deepflow}; (b) Scene flow methods PRSM \cite{Vogel2015IJCV}; and (c) Non-sequential alignment of partial surfaces 4DMatch \cite{Mustafa16ECCV} (requires a prior 3D mesh of the object as input for 4D reconstruction). 
The key-frames of sequence are coloured and the colour is propagated using dense flow from the joint optimisation throughout the sequence. The red regions in 2D dense flow in Figure \ref{fig:flow} are the regions for which reliable correspondences are not found. This demonstrates improved performance using the proposed method. The colours in the 4D alignment in Figure \ref{fig:4Deval} are not reliably propagated by DCFlow for limbs.

We also compare the silhouette overlap error ($S_e$) across frames, key-frames and views to evaluate long-term temporal coherence in Table \ref{soe} for all datasets. This is defined as $S_e = \frac{1}{N_{v} N_{k} N_{f}^{i}}\sum_{i = 1}^{N_{k}}\sum_{j = 1}^{N_{f}^{i}}\sum_{c = 1}^{N_{v}} \frac{\text{Area of intersection}}{\text{Area of semantic segmentation}}$. Dense flow in time is used to obtain the propagated mask for each image. The propagated mask is overlapped with semantic segmentation at each time instant to evaluate the accuracy of the propagated mask. The lower the $S_e$ the better. Our approach gives the lowest error demonstrating higher accuracy compared to the state-of-the-art methods.
\vspace{-0.1cm}
%
\subsection{Ablation study on Equation \ref{eq:costfunction}}
\label{sec:ablation}
\vspace{-0.15cm}
We perform an ablation study on Equation \ref{eq:costfunction}, such that we remove motion $E_{f}, E_{r}$, pose $E_{p}$ and semantic $E_{sem}$ constraints from the equation, defining $P_{M} = E- E_{f} - E_{r}, P_{P} = E- E_{p}, P_{PM} = E- E_{f} - E_{r} - E_{p}, P_{S} = E- E_{sem}$ and $P_{PS} = E- E_{sem} - E_{p}$. Reconstruction, flow and semantic segmentation is obtained with removed constraints, and the results are shown in Tables \ref{reconstructCompare}, \ref{soe} and \ref{segTable} respectively. The proposed approach gives best performance with joint pose, motion and semantic constraints. 
%
\vspace{-0.1cm}
\subsection{Limitations}
\vspace{-0.15cm}
Gross errors in initial semantic instance segmentation and 3D pose estimation lead to degradation in the quality of results (e.g. the cars in Juggler2 - Figure \ref{fig:segment}). Although 3D human pose helps in robust 4D reconstruction of interacting people in dynamic scenes, current 3D pose estimation is unreliable for highly crowded environments resulting in degradation of the proposed approach.
%
\section{Conclusions}
\label{sec:conclude}
\vspace{-0.15cm}
This paper introduced the first method for unsupervised 4D dynamic scene understanding from multi-view video.
A novel joint flow, reconstruction and semantic instance segmentation estimation framework is introduced exploiting 2D/3D human-pose, motion, semantic, shape and appearance information in space and time.  
Ablation study on the joint optimisation demonstrates the effectiveness of the proposed scene understanding framework for general scenes with multiple interacting people.
The semantic, motion and depth information per view is fused spatially across views for 4D semantically and temporally coherent scene understanding.
Extensive evaluation against state-of-the-art methods on a variety of complex indoor and outdoor datasets with large non-rigid deformations demonstrates a significant improvement in the accuracy in semantic segmentation, reconstruction, motion estimation and 4D alignment. 
{\small
	\bibliographystyle{ieee}
	\bibliography{egbib}

\begin{thebibliography}{10}\itemsep=-1pt

\bibitem{4DInria}
4d repository, http://4drepository.inrialpes.fr/.
\newblock In {\em Institut national de recherche en informatique et en
  automatique (INRIA) Rhone Alpes}.

\bibitem{cvssp3d}
Multiview video repository, http://cvssp.org/data/cvssp3d/.
\newblock In {\em Centre for Vision Speech and Signal Processing, University of
  Surrey, UK}.

\bibitem{badrinarayanan2015segnet}
V.~Badrinarayanan, A.~Kendall, and R.~Cipolla.
\newblock Segnet: A deep convolutional encoder-decoder architecture for image
  segmentation.
\newblock {\em TPAMI}, 2017.

\bibitem{UnstructuredVBR10}
L.~Ballan, G.~J. Brostow, J.~Puwein, and M.~Pollefeys.
\newblock Unstructured video-based rendering: Interactive exploration of
  casually captured videos.
\newblock {\em ACM Trans. Graph.}, 29(4):1--11, 2010.

\bibitem{Basha2013}
T.~Basha, Y.~Moses, and N.~Kiryati.
\newblock Multi-view scene flow estimation: A view centered variational
  approach.
\newblock In {\em CVPR}, pages 1506--1513, 2010.

\bibitem{Kolmogorov04}
Y.~Boykov and V.~Kolmogorov.
\newblock An experimental comparison of min-cut/max- flow algorithms for energy
  minimization in vision.
\newblock {\em TPAMI}, 26(11):1124--1137, 2004.

\bibitem{Boykov01}
Y.~Boykov, O.~Veksler, and R.~Zabih.
\newblock Fast approximate energy minimization via graph cuts.
\newblock {\em TPAMI}, 23(11):1222--1239, 2001.

\bibitem{cao2017realtime}
Z.~Cao, T.~Simon, S.-E. Wei, and Y.~Sheikh.
\newblock Realtime multi-person 2d pose estimation using part affinity fields.
\newblock In {\em CVPR}, 2017.

\bibitem{deeplabv3}
L.~Chen, Y.~Zhu, G.~Papandreou, F.~Schroff, and H.~Adam.
\newblock Encoder-decoder with atrous separable convolution for semantic image
  segmentation.
\newblock {\em CoRR}, abs/1802.02611, 2018.

\bibitem{ChenPK0Y16}
L.~C. Chen, G.~Papandreou, I.~Kokkinos, K.~Murphy, and A.~L. Yuille.
\newblock Deeplab: Semantic image segmentation with deep convolutional nets,
  atrous convolution, and fully connected crfs.
\newblock {\em CoRR}, abs/1606.00915, 2016.

\bibitem{chiu2013coseg}
W.-C. Chiu and M.~Fritz.
\newblock Multi-class video co-segmentation with a generative multi-video
  model.
\newblock In {\em CVPR}, 2013.

\bibitem{Djelouah16}
A.~Djelouah, J.-S. Franco, E.~Boyer, P.~P{\'e}rez, and G.~Drettakis.
\newblock {Cotemporal Multi-View Video Segmentation}.
\newblock In {\em 3DV}, 2016.

\bibitem{EngelmannGCPR16}
F.~Engelmann, J.~St\"uckler, and B.~Leibe.
\newblock Joint object pose estimation and shape reconstruction in urban street
  scenes using {3D} shape priors.
\newblock In {\em GCPR}, 2016.

\bibitem{pascal-voc-2012}
M.~Everingham, L.~Van~Gool, C.~K.~I. Williams, J.~Winn, and A.~Zisserman.
\newblock The {PASCAL} {V}isual {O}bject {C}lasses {C}hallenge 2012 {(VOC2012)}
  {R}esults.
\newblock
  http://www.pascal-network.org/challenges/VOC/voc2012/workshop/index.html.

\bibitem{FarabetCNL13}
C.~Farabet, C.~Couprie, L.~Najman, and Y.~LeCun.
\newblock Learning hierarchical features for scene labeling.
\newblock {\em TPAMI}, 35(8):1915--1929, 2013.

\bibitem{Floros12}
G.~Floros and B.~Leibe.
\newblock Joint 2d-3d temporally consistent semantic segmentation of street
  scenes.
\newblock In {\em CVPR}, pages 2823--2830, 2012.

\bibitem{Guillemaut2010}
J.~Y. Guillemaut and A.~Hilton.
\newblock {Joint Multi-Layer Segmentation and Reconstruction for Free-Viewpoint
  Video Applications}.
\newblock {\em IJCV}, 93:73--100, 2010.

\bibitem{Gupta2014}
S.~Gupta, R.~Girshick, P.~Arbel{\'a}ez, and J.~Malik.
\newblock {\em Learning Rich Features from RGB-D Images for Object Detection
  and Segmentation}, pages 345--360.
\newblock 2014.

\bibitem{HanePAMI2016}
C.~Hane, C.~Zach, A.~Cohen, and M.~Pollefeys.
\newblock Dense semantic 3d reconstruction.
\newblock {\em TPAMI}, page~1, 2016.

\bibitem{HariharanAGM15}
B.~Hariharan, P.~A. Arbeláez, R.~B. Girshick, and J.~Malik.
\newblock Hypercolumns for object segmentation and fine-grained localization.
\newblock In {\em CVPR}, pages 447--456, 2015.

\bibitem{2017maskrcnn}
K.~He, G.~Gkioxari, P.~Doll\'{a}r, and R.~Girshick.
\newblock {Mask R-CNN}.
\newblock In {\em ICCV}, 2017.

\bibitem{MuVS3DV2017}
Y.~Huang, F.~Bogo, C.~Lassner, A.~Kanazawa, P.~V. Gehler, J.~Romero, I.~Akhter,
  and M.~J. Black.
\newblock Towards accurate marker-less human shape and pose estimation over
  time.
\newblock In {\em 3DV}, 2017.

\bibitem{h36m_pami}
C.~Ionescu, D.~Papava, V.~Olaru, and C.~Sminchisescu.
\newblock Human3.6m: Large scale datasets and predictive methods for 3d human
  sensing in natural environments.
\newblock {\em TPAMI}, 36(7):1325--1339, jul 2014.

\bibitem{Kazhdan2006}
M.~Kazhdan, M.~Bolitho, and H.~Hoppe.
\newblock Poisson surface reconstruction.
\newblock In {\em Eurographics Symposium on Geometry Processing}, pages 61--70,
  2006.

\bibitem{KendallGC17}
A.~Kendall, Y.~Gal, and R.~Cipolla.
\newblock Multi-task learning using uncertainty to weigh losses for scene
  geometry and semantics.
\newblock {\em CoRR}, abs/1705.07115, 2017.

\bibitem{Kim2012}
H.~Kim, J.~Guillemaut, T.~Takai, M.~Sarim, and A.~Hilton.
\newblock {Outdoor Dynamic 3-D Scene Reconstruction}.
\newblock {\em T-CSVT}, 22(11):1611--1622, 2012.

\bibitem{Abhijit14}
A.~Kundu, Y.~Li, F.~Dellaert, F.~Li, and J.~M. Rehg.
\newblock Joint semantic segmentation and 3d reconstruction from monocular
  video.
\newblock In {\em ECCV}, volume 8694, pages 703--718, 2014.

\bibitem{Kundu2016CVPR}
A.~Kundu, V.~Vineet, and V.~Koltun.
\newblock Feature space optimization for semantic video segmentation.
\newblock In {\em CVPR}, pages 3168--3175, 2016.

\bibitem{langguth-2016-smvs}
F.~Langguth, K.~Sunkavalli, S.~Hadap, and M.~Goesele.
\newblock Shading-aware multi-view stereo.
\newblock In {\em ECCV}, 2016.

\bibitem{Larsen07}
E.~Larsen, P.~Mordohai, M.~Pollefeys, and H.~Fuchs.
\newblock Temporally consistent reconstruction from multiple video streams
  using enhanced belief propagation.
\newblock In {\em ICCV}, pages 1--8, 2007.

\bibitem{LinMBHPRDZ14}
T.-Y. Lin, M.~Maire, S.~J. Belongie, L.~D. Bourdev, R.~B. Girshick, J.~Hays,
  P.~Perona, D.~Ramanan, P.~Doll{\'{a}}r, and C.~L. Zitnick.
\newblock Microsoft {COCO:} common objects in context.
\newblock {\em CoRR}, abs/1405.0312, 2014.

\bibitem{long_shelhamer_fcn}
J.~Long, E.~Shelhamer, and T.~Darrell.
\newblock Fully convolutional networks for semantic segmentation.
\newblock In {\em CVPR}, 2015.

\bibitem{Lowe04}
D.~G. Lowe.
\newblock Distinctive image features from scale-invariant keypoints.
\newblock {\em IJCV}, 60:91--110, 2004.

\bibitem{Luo2015}
B.~Luo, H.~Li, T.~Song, and C.~Huang.
\newblock Object segmentation from long video sequences.
\newblock In {\em ACM Multimedia}, pages 1187--1190, 2015.

\bibitem{MostajabiYS15}
M.~Mostajabi, P.~Yadollahpour, and G.~Shakhnarovich.
\newblock Feedforward semantic segmentation with zoom-out features.
\newblock In {\em CVPR}, pages 3376--3385, 2015.

\bibitem{MustafaCVPR17}
A.~Mustafa and A.~Hilton.
\newblock Semantically coherent co-segmentation and reconstruction of dynamic
  scenes.
\newblock In {\em CVPR}, 2017.

\bibitem{Mustafa16CVPR}
A.~Mustafa, H.~Kim, J.-Y. Guillemaut, and A.~Hilton.
\newblock Temporally coherent 4d reconstruction of complex dynamic scenes.
\newblock In {\em CVPR}, 2016.

\bibitem{Mustafa16ECCV}
A.~Mustafa, H.~Kim, and A.~Hilton.
\newblock 4d match trees for non-rigid surface alignment.
\newblock In {\em ECCV}, 2016.

\bibitem{Mustafa3DV17}
A.~Mustafa, M.~Volino, J.-Y. Guillemaut, and A.~Hilton.
\newblock 4d temporally coherent light-field video.
\newblock In {\em 3DV}, 2017.

\bibitem{Newcombe2015DynamicFusionRA}
R.~A. Newcombe, D.~Fox, and S.~M. Seitz.
\newblock Dynamicfusion: Reconstruction and tracking of non-rigid scenes in
  real-time.
\newblock {\em CVPR}, pages 343--352, 2015.

\bibitem{Agapito2012}
A.~Roussos, C.~Russell, R.~Garg, and L.~Agapito.
\newblock Dense multibody motion estimation and reconstruction from a handheld
  camera.
\newblock In {\em ISMAR}, 2012.

\bibitem{sevillaCVPR2016}
L.~Sevilla-Lara, D.~Sun, V.~Jampani, and M.~J. Black.
\newblock Optical flow with semantic segmentation and localized layers.
\newblock In {\em CVPR}, pages 3889--3898, 2016.

\bibitem{Sorkine2007}
O.~Sorkine and M.~Alexa.
\newblock As-rigid-as-possible surface modeling.
\newblock In {\em SGP}, pages 109--116, 2007.

\bibitem{Taniai18}
T.~Taniai, Y.~Matsushita, Y.~Sato, and T.~Naemura.
\newblock {Continuous 3D Label Stereo Matching using Local Expansion Moves}.
\newblock {\em {TPAMI}}, 40(11):2725--2739, 2018.

\bibitem{Tao2012}
M.~W. Tao, J.~Bai, P.~Kohli, and S.~Paris.
\newblock Simpleflow: A non-iterative, sublinear optical flow algorithm.
\newblock {\em Computer Graphics Forum (Eurographics 2012)}, 31(2), May 2012.

\bibitem{Tome_2017_CVPR}
D.~Tome, C.~Russell, and L.~Agapito.
\newblock Lifting from the deep: Convolutional 3d pose estimation from a single
  image.
\newblock In {\em CVPR}, July 2017.

\bibitem{tome3DV2018}
D.~Tom{\`{e}}, M.~Toso, L.~Agapito, and C.~Russell.
\newblock Rethinking pose in 3d: Multi-stage refinement and recovery for
  markerless motion capture.
\newblock In {\em 3DV}, 2018.

\bibitem{TsaiZ016}
Y.-H. Tsai, G.Zhong, and M.-H. Yang.
\newblock Semantic co-segmentation in videos.
\newblock In {\em ECCV}, pages 760--775, 2016.

\bibitem{Ulusoy2017CVPR}
A.~O. Ulusoy, M.~J. Black, and A.~Geiger.
\newblock Semantic multi-view stereo: Jointly estimating objects and voxels.
\newblock In {\em CVPR}, 2017.

\bibitem{vineet2015icra}
V.~Vineet, O.~Miksik, M.~Lidegaard, M.~Nie{\ss}ner, S.~Golodetz, V.~A.
  Prisacariu, O.~K\"ahler, D.~W. Murray, S.~Izadi, P.~Perez, and P.~H.~S. Torr.
\newblock Incremental dense semantic stereo fusion for large-scale semantic
  scene reconstruction.
\newblock In {\em ICRA}, 2015.

\bibitem{Vlasic2008}
D.~Vlasic, I.~Baran, W.~Matusik, and J.~Popovi\'{c}.
\newblock Articulated mesh animation from multi-view silhouettes.
\newblock {\em ACM Trans. Graph.}, 27(3), Aug. 2008.

\bibitem{Vogel2015IJCV}
C.~Vogel, K.~Schindler, and S.~Roth.
\newblock 3d scene flow estimation with a piecewise rigid scene model.
\newblock pages 1--28, 2015.

\bibitem{Wedel2011}
A.~Wedel, T.~Brox, T.~Vaudrey, C.~Rabe, U.~Franke, and D.~Cremers.
\newblock Stereoscopic scene flow computation for 3d motion understanding.
\newblock {\em IJCV}, 95(1):29--51, 2011.

\bibitem{deepflow}
P.~Weinzaepfel, J.~Revaud, Z.~Harchaoui, and C.~Schmid.
\newblock Deepflow: Large displacement optical flow with deep matching.
\newblock In {\em ICCV}, pages 1385--1392, 2013.

\bibitem{Xia2017CVPR}
F.~Xia, P.~Wang, X.~Chen, and A.~L. Yuille.
\newblock Joint multi-person pose estimation and semantic part segmentation.
\newblock In {\em CVPR}, 2017.

\bibitem{Xie2016CVPR}
J.~Xie, M.~Kiefel, M.-T. Sun, and A.~Geiger.
\newblock Semantic instance annotation of street scenes by 3d to 2d label
  transfer.
\newblock In {\em CVPR}, 2016.

\bibitem{XRK2017}
J.~Xu, R.~Ranftl, and V.~Koltun.
\newblock {Accurate Optical Flow via Direct Cost Volume Processing}.
\newblock In {\em CVPR}, 2017.

\bibitem{Zanfir_2015_ICCV}
A.~Zanfir and C.~Sminchisescu.
\newblock Large displacement 3d scene flow with occlusion reasoning.
\newblock In {\em ICCV}, 2015.

\bibitem{zhao2017pspnet}
H.~Zhao, J.~Shi, X.~Qi, X.~Wang, and J.~Jia.
\newblock Pyramid scene parsing network.
\newblock In {\em CVPR}, 2017.

\bibitem{crfasrnn_iccv2015}
S.~Zheng, S.~Jayasumana, B.~Romera-Paredes, V.~Vineet, Z.~Su, D.~Du, C.~Huang,
  and P.~Torr.
\newblock Conditional random fields as recurrent neural networks.
\newblock In {\em ICCV}, 2015.

\end{thebibliography}
}
\end{document}